\documentclass[journal]{IEEEtran}

\usepackage{amsmath,amssymb,amsfonts}
\usepackage{graphicx}
\usepackage{cite}
\usepackage{url}
\usepackage{hyperref}
\usepackage{booktabs}
\usepackage{caption}

\usepackage{subcaption}
\usepackage{xcolor}
\usepackage{tcolorbox}
\usepackage{fancyhdr}
\usepackage{titlesec}
\usepackage{float}
\usepackage{tabularx}
\usepackage{siunitx}
\definecolor{ieeeblue}{RGB}{0,102,153}

\fancypagestyle{header}{
    \fancyhf{}
    
    \fancyhead[C]{\colorbox{ieeeblue}{\parbox{\columnwidth}{\centering \textcolor{white}{\textbf{VEHICULAR NETWORKING}}}}}
}

\begin{document}

\thispagestyle{header}

\title{\huge \textbf{Strengthening Multi-Robot Systems for SAR: Co-Designing Robotics and Communication Towards 6G}}

\author{\IEEEauthorblockN{\large Juan Bravo-Arrabal, Ricardo Vázquez-Martín, J.J. Fernández-Lozano, Alfonso García-Cerezo}\\
\IEEEauthorblockA{\textit{} \\
\textit{Institute for Mechatronics Engineering and Cyber-physical Systems (IMECH.UMA)}} \\
\textit{UMA Robotics and Mechatronics Group \\
\textit{University of Malaga (UMA), Spain} \\
Email: jbravo@uma.com}}

\maketitle

\begin{tcolorbox}[colback=white, colframe=white, sharp corners, boxrule=0mm]
\noindent \textbf{\textcolor{ieeeblue}{Abstract}}

This paper presents field-tested use cases from Search and Rescue (SAR) missions, highlighting the co-design of mobile robots and communication systems to support Edge-Cloud architectures based on 5G Standalone (SA). The main goal is to contribute to the effective cooperation of multiple robots and first responders. Our field experience includes the development of Hybrid Wireless Sensor Networks (H-WSNs) for risk and victim detection, smartphones integrated into the Robot Operating System (ROS) as Edge devices for mission requests and path planning, real-time Simultaneous Localization and Mapping (SLAM) via Multi-Access Edge Computing (MEC), and implementation of Uncrewed Ground Vehicles (UGVs) for victim evacuation in different navigation modes. These experiments, conducted in collaboration with actual first responders, underscore the need for intelligent network resource management, balancing low-latency and high-bandwidth demands. Network slicing is key to ensuring critical emergency services are performed despite challenging communication conditions. The paper identifies architectural needs, lessons learned, and challenges to be addressed by 6G technologies to enhance emergency response capabilities.

\end{tcolorbox}


\titleformat{\section}{\color{ieeeblue}\large\bfseries}{\thesection}{1em}{}

\section{Introduction}
\label{sec:intro}

In recent decades, new technologies have been increasingly used to support and assist in rescue operations improving safety and efficiency in emergency response. In this context, robots play a key role in collecting data from disaster environments, searching for victims, conducting structural inspections, providing medical assistance and intervention, recovering victims, extending rescue teams, and performing other logistical tasks \cite{Murphy2016}. For safety reasons, teleoperation has been used in robot deployments in disaster situations. However, achieving full autonomy remains a significant challenge due to the complexity of emergency scenarios. As a result, semi-autonomous tasks are often preferred, where robustness and coordination among multiple multimodal robots and human teams are the main objectives \cite{Delmerico2019}. In addition, creating an IoT–edge–cloud continuum remains challenging but could greatly benefit SAR operations \cite{Militano2023}. 

Semi-autonomous robotic tasks that require interactions among robots, the environment, and response teams depend on high-quality wireless communication systems for effective collaboration and coordination toward shared goals. 5G technology delivers reliable wireless connectivity with low latency essential for demanding applications such as remote sensing and mapping, while also enabling Edge computing capabilities. Furthermore, 5G features can provide a resilient network infrastructure, deploying crowdcells in disaster areas where catastrophic events usually destroy communication systems.   

This article summarizes our experience applying communication and robotic technologies to emergency response operations, conducted during outdoor field exercises designed with high authenticity by and for actual search-and-rescue (SAR) professionals. The reported use cases can be categorized into two groups: those that extend first-responder capabilities, such as information gathering, and those that complement or support physical activities, such as casualty extraction. For each use case, we describe the mission objectives, participating stakeholders, robotic systems and end-users, the specific contribution of robots and communication technologies, and the observed operational outcomes and challenges. The article concludes with an analysis of identified limitations, key lessons learned, and recommendations for enhancing communication reliability in emergency scenarios.

The remainder of this paper is organized as follows. Section \ref{sec:emergency} describes our approach to communication and robotic systems for disaster robotics applications, in Section \ref{sec:uc} we summarize use cases in different emergency response missions in high-fidelity field experiments. In Section \ref{sec:discussion} we evaluate the results of our field experiments, outline problems, limitations, lessons learned, and finally provide future perspectives. Finally, Section \ref{sec:concl} offers conclusions.

\label{sec:emergency}

\section{An approach to communication and robotic systems for emergency missions}

Collaboration with actual first responders is a cornerstone in our approach to emergency robotics. This partnership grounds technological development in real operational needs by providing direct insights from fire departments, police, and other emergency services. A key driver is the annual Workshop on Emergencies, called JEMERG\footnote{\scriptsize JEMERG: Jornadas de Seguridad, Emergencias y Catástrofes.}, organized by the Chair for Security and Emergencies of the University of Malaga (UMA), led by Professor Jesús Miranda. Since its first edition in 2006, JEMERG has become a reference event, gathering around 200 people in a multidisciplinary field exercise conducted in a realistic setting. These exercises have effectively incorporated human factors into our approach, considering operational procedures, workload management, stress-elements, and actual needs, identifying practical limitations not visible in laboratory settings.

As a result, our approach relies on the development of use cases focused on meeting the specific needs of actual first responders, considering organizational and operational integration of the technical solutions into response teams. The resulting development loop drives continuous improvement, ensuring that our robotic solutions remain connected to the practical realities of emergency response, facilitating also later adoption by these end-users.

Another key element of our approach derived directly from our collaboration with first responders. Emergency operations depend critically on reliable information, which must be up to date. The first step to meet this need was to use Wireless Sensor Networks (WSN). Static sensors remain fixed at strategic locations collecting environmental data (temperature, gas, etc.) to detect risks like fires and earthquakes. Hybrid WSN (H-WSN) architectures based on technologies like LoRa and ZigBee, incorporating static and mobile nodes, extend coverage, and adapt to dynamic mission requirements. Mobile nodes attached to robots, rescue dogs, and crewed vehicles move throughout the environment to update environmental information and to adapt to changing conditions. Another relevant use is victim detection. Using Bluetooth Low Energy (BLE) technology, the robots can detect potential survivors through their electronic devices. When a robot's scanner picks up a signal, it immediately logs the location via GNNS and shares this information through the network, allowing rescue teams to prioritize areas with the highest probability of finding survivors.

\begin{figure}[t]
     \centering     
     \includegraphics[trim={85 70 53 92},clip, width=\columnwidth]{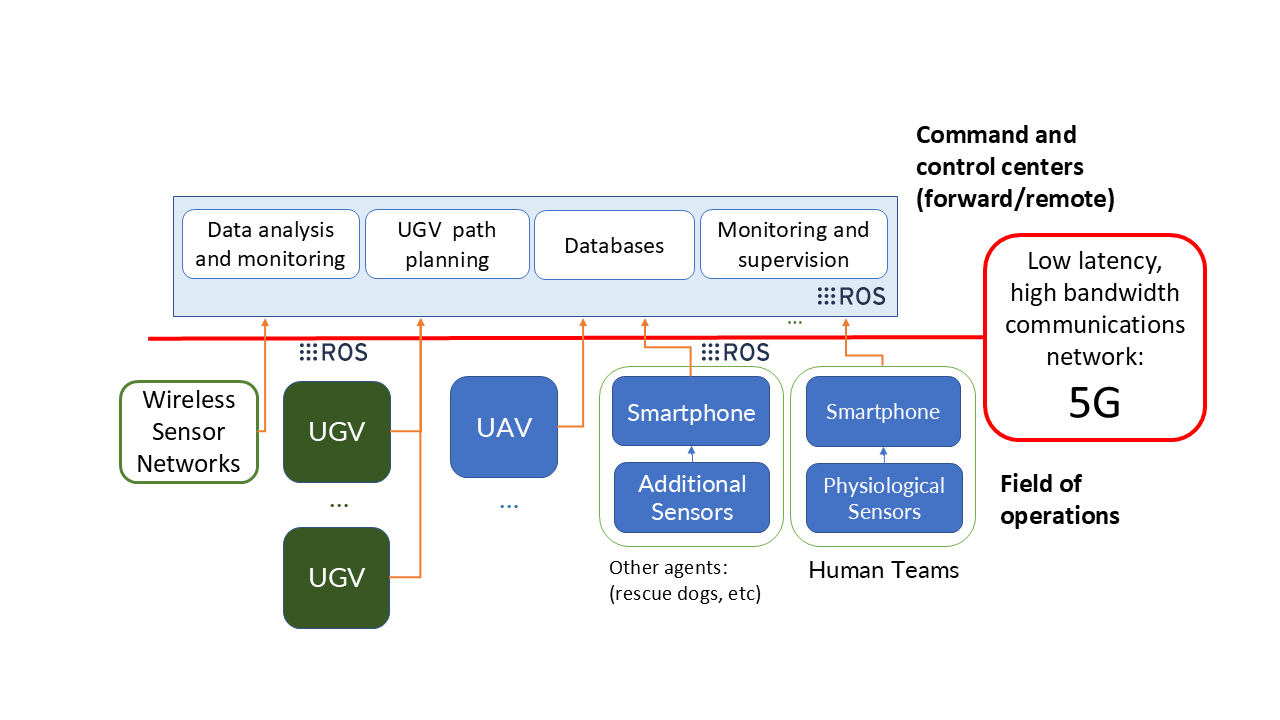}
     \caption{Scheme for a multi-robot system in SAR coordination}
     \label{fig:general-scheme}
\end{figure}

This use of mobile robots integrated into a WSN highlighted the potential of connected systems. The next step was the incorporation of high-bandwidth, low-latency communication systems like 5G. Within a pilot project funded by the Spanish Government and led by Vodafone and Huawei, the group developed and tested 5G-integrated solutions in realistic exercises. 5G enabled extensive information sharing and new use cases but also demanded a holistic approach to integrate diverse elements: ground and aerial robots, crewed vehicles, WSNs, human responders, and control centers. The result was the Internet of Cooperative Agents architecture (X-IoCA) \cite{Bravo-Arrabal2021}, which integrates all these components and introduces features like Multi-access Edge Computing (MEC) centers for efficient data sharing, computation, and coordination in complex environments. Figure~\ref{fig:general-scheme} reflects our current approach to emergency operations as a collective mission performed by a heterogeneous team of agents exchanging information. Communications are now central, enabling use cases like remote casualty extraction, edge-based mapping, real-time responder stress monitoring \cite{Vera-Ortega2022}, and cooperative robotic missions.

\section{Use cases in connected disaster robotics}
\label{sec:uc}

The Robotics and Mechatronics Group, together with the Institute of Mechatronics of the University of Malaga (IMECH.UMA), have created the Laboratory and Area for Experimentation in New Technologies for Catastrophes (LAENTIEC). This initiative was led by Professor Alfonso García-Cerezo, as part of his long-standing commitment to advancing technological solutions for emergency response and disaster management. This facility is dedicated to research and innovation in the field of disaster robotics, with the aim of providing society with effective tools to save lives in a wide range of emergency scenarios, including natural catastrophes (such as earthquakes, floods, or volcanic eruptions) and human-made disasters (such as building collapses and armed conflicts).

LAENTIEC comprises a laboratory facility adjacent to a terrain covering approximately \SI{90000}{\meter\squared}, characterized by varied orography, a river bed, a dual-tunnel stretching \SI{140}{\meter}, and differing degrees of accessibility across its experimental zones. LAENTIEC hosts JEMERG, focused on disaster scenarios, serving as a platform for testing and validating emerging technologies under realistic and challenging conditions.

The following use cases demonstrate the integration of mobile robots and first responders as cooperative agents in disaster scenarios, highlighting the importance of tailored application design and robust communication strategies.

\subsection{Perception and localization via Hybrid WSN}

Mobile nodes can function as both Concentrator Nodes (CNs) and End Devices (EDs), i.e., Edge devices. Any mobile SAR agent (e.g., firefighting brigades, military units, or mobile robots) has a certain payload capacity to deploy WSN nodes, thereby extending network coverage and enhancing the efficiency of risk detection and victim localization.

\begin{figure*}[t]
     \centering     \includegraphics[width=.85\textwidth]{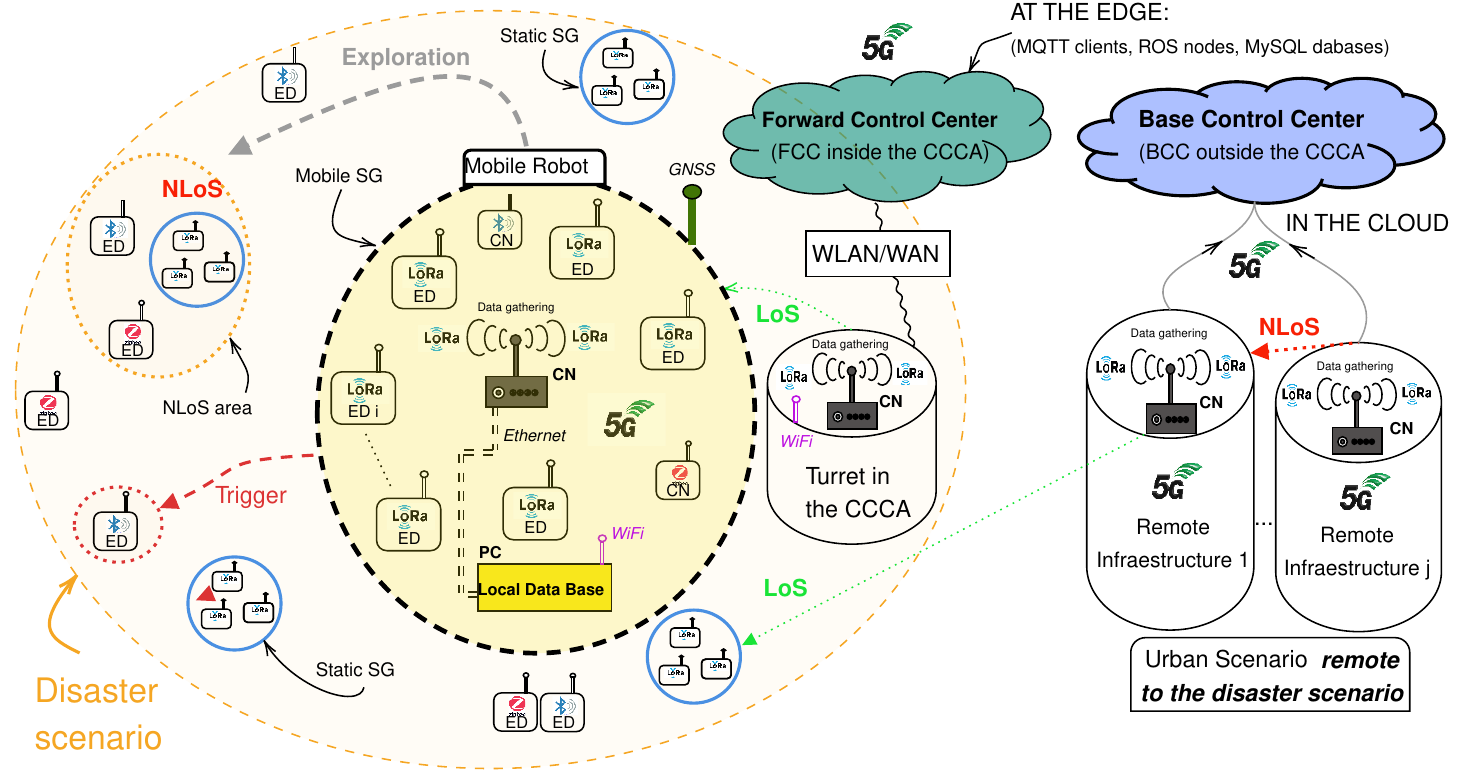}
     \caption{Framework for mobile robot interaction in SAR scenarios, including H-WSNs.}
     \label{fig:hwsn-case}
\end{figure*}

Figure \ref{fig:hwsn-case} depicts a framework for mobile robot interaction with its operational environment through various H-WSN, including Bluetooth Low Energy (BLE), LoRa, and Zigbee.

The robot is integrated into all these networks by carrying and transporting nodes of each technology as it navigates throughout the environment. Additionally, static EDs are deployed in Sensory Groups (SGs) by human agents in areas of interest, including those without Line of Sight (LoS).

These SGs can measure environmental variables and detect relevant gradients (such as temperature, gas, humidity, or noise), actively contributing to the construction of shared situational awareness. Each SG is equipped with a GNSS sensor that provides its position, which is transmitted via LoRa to the Forward Control Center (FCC). If the readings from any SG exceed a predefined threshold, an alarm is triggered and sent to the FCC. The FCC then computes a path plan for the UGV closest to the location reported by the corresponding SG. In this way, robots provide mobility and access to areas that are difficult or hazardous for human teams, acting as sensing units, Edge-processing nodes, communication relays, or coverage extenders \cite{Bravo-Arrabal2023}.

Within this hybrid architecture, mesh networks using Zigbee and WiFi have been explored to enhance distributed environmental perception and extend communication coverage. UGVs and UAVs are equipped with embedded Edge devices that store data locally and synchronize with a central database only when network connectivity is available, allowing delayed but reliable data delivery to the Command, Control, and Communication Area (CCCA). Notably, some mobile robots serve as data Mobile Ubiquitous LAN Extensions (MULE): they collect data in disconnected areas and offload it later when within communication range \cite{Bravo-Arrabal2022}. This strategy ensures system robustness even in scenarios with intermittent or no infrastructure-based connectivity.

These SGs can measure environmental variables and detect relevant gradients (such as temperature, gas, humidity, or noise), actively contributing to the construction of shared situational awareness. Each SG is equipped with a GNSS sensor that provides its position, which is transmitted via LoRa to the Forward Control Center (FCC). If the readings from any SG exceed a predefined threshold, an alarm is triggered and sent to the FCC. The FCC then computes a path plan for the UGV closest to the location reported by the corresponding SG. In this way, robots provide mobility and access to areas that are difficult or hazardous for human teams, acting as sensing units, Edge-processing nodes, communication relays, or coverage extenders \cite{Bravo-Arrabal2023}.

Within this hybrid architecture, mesh networks using Zigbee and WiFi have been explored to enhance distributed environmental perception and extend communication coverage. UGVs and UAVs are equipped with embedded Edge devices that store data locally and synchronize with a central database only when network connectivity is available, allowing delayed but reliable data delivery to the Command, Control, and Communication Area (CCCA). Notably, some mobile robots serve as data Mobile Ubiquitous LAN Extensions (MULE): they collect data in disconnected areas and offload it later when within communication range \cite{Bravo-Arrabal2022}. This strategy ensures system robustness even in scenarios with intermittent or no infrastructure-based connectivity.

In this setup, a modified version of the LoRaWAN protocol was evaluated, allowing a single packet—transmitted by a distributed SG—to be received simultaneously by multiple gateways acting as CNs \cite{Bravo-Arrabal2022}. This feature enables localization services over LoRa, even when relying on remote CNs installed on urban rooftops \cite{bravo2021development}. These remote urban gateways can be connected to the Internet via 5G, providing Cloud services to the disaster area through the LoRaWAN network, ensuring data integration even under NLoS conditions.

Finally, victim detection has also been explored through passive scanning of BLE signals from both UGV and UAV platforms \cite{Cantizani-Estepa2022}, analyzing parameters such as Received Signal Strength Indicator (RSSI) and Signal to Noise Ratio (SNR), device manufacturer data, and the dynamic or static nature of detected Medium Access Control (MAC) addresses. These indicators help infer the presence of potential victims in the field.

Results from our H-WSN implementations demonstrate that the true value of connected networks lies not in high bandwidth but in low latency, resilient connectivity, and the ability to enable efficient distributed coordination. In this regard, adopting 5G —and its evolution toward 6G— is crucial for fulfilling the requirements of synchronization, scalability, and robustness demanded by effective H-WSNs in dynamic and challenging environments.

\subsection{Casualty extraction using smartphones as Edge devices}

\begin{figure}[t]
    \centering

    \begin{subfigure}[b]{0.8\columnwidth}
        \includegraphics[width=\linewidth]{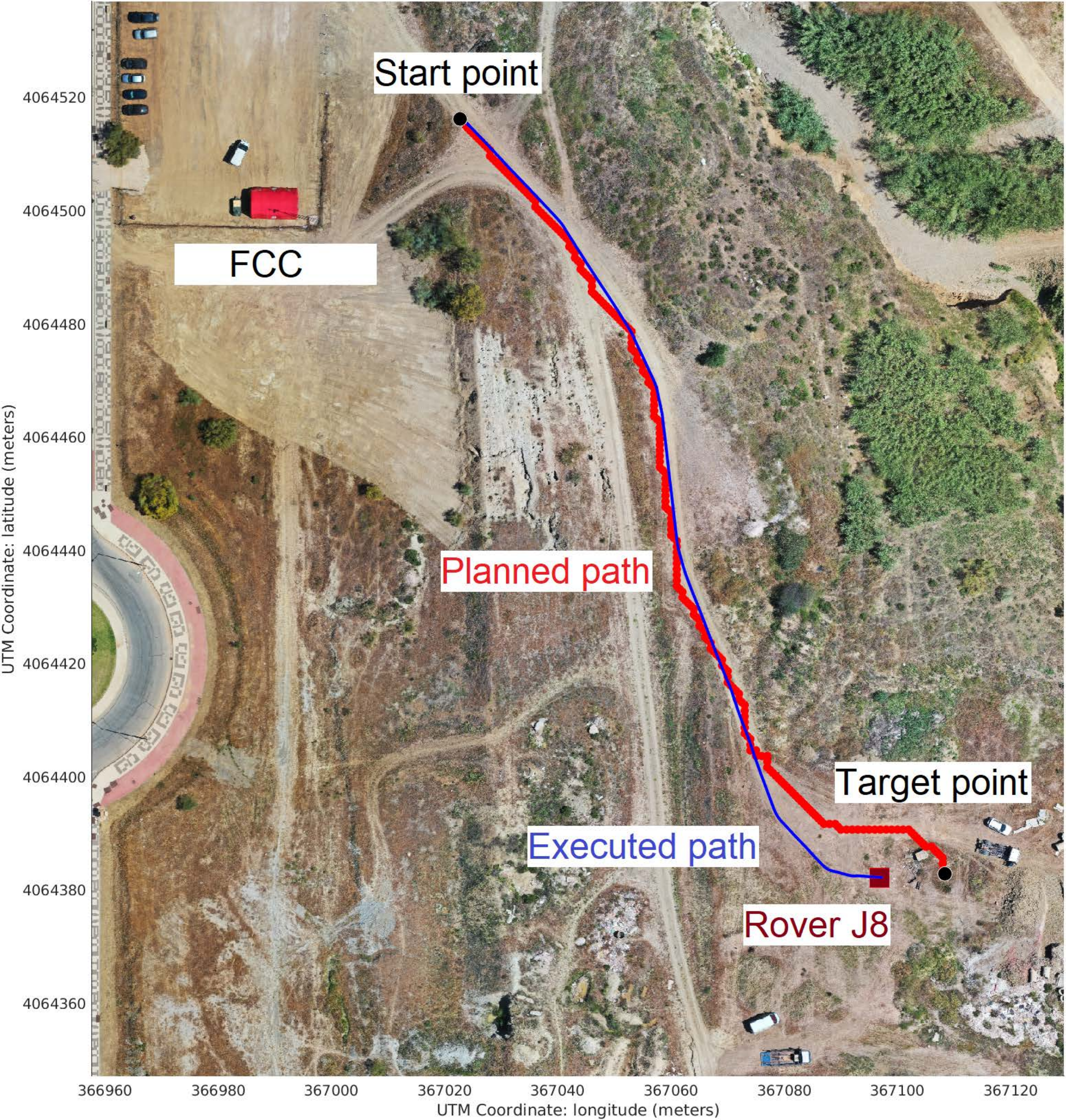}
        \caption{}
        \label{fig:requests-case-a}
    \end{subfigure}
    
    \begin{subfigure}[b]{0.8\columnwidth}
        \includegraphics[width=\linewidth]{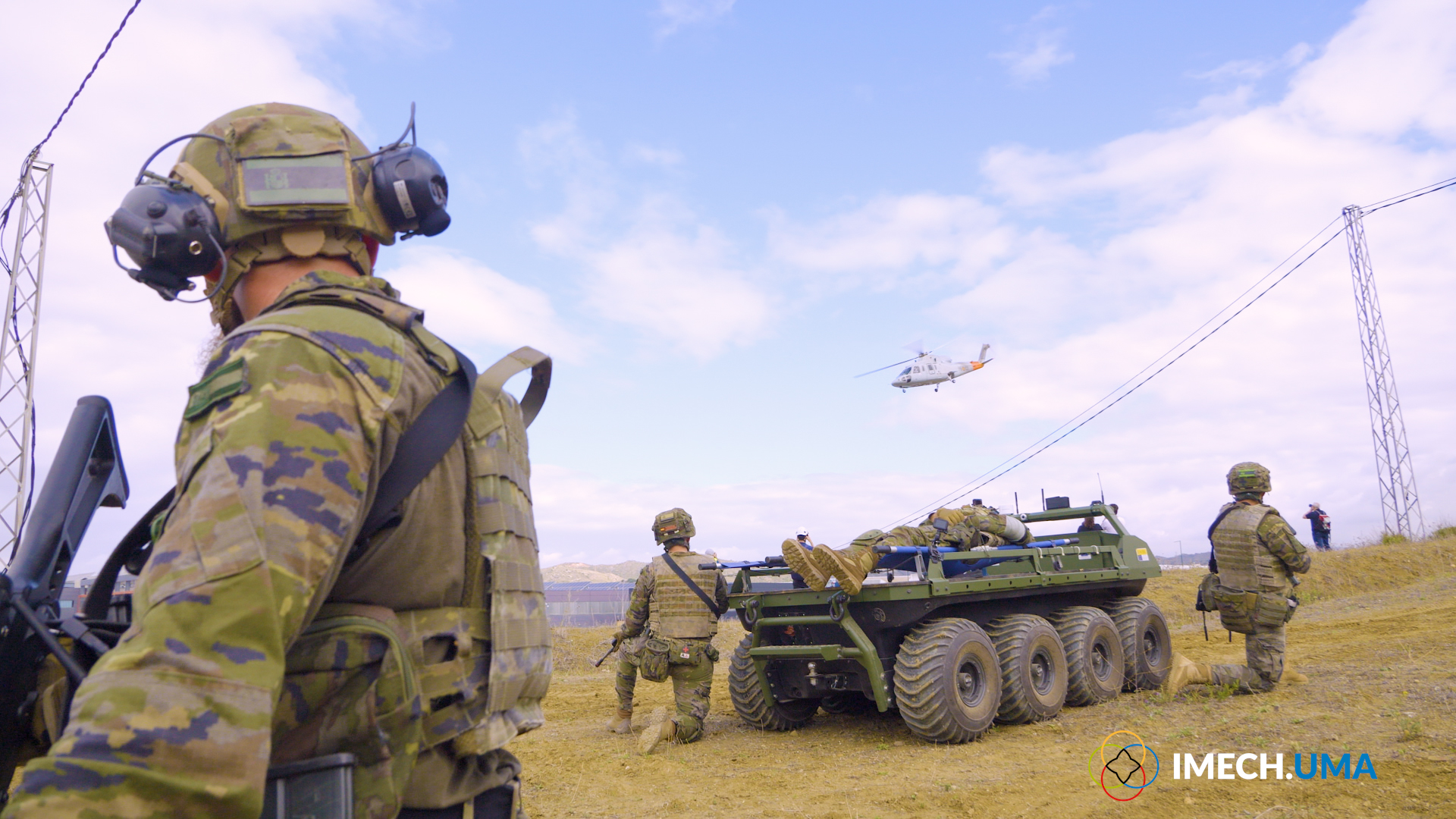}
        \caption{}
        \label{fig:requests-case-b}
    \end{subfigure}

    \caption{Autonomous victim evacuation: (a) path-planning execution; (b) Rover J8 in position, waiting for the helicopter.}
    \label{fig:requests-case}
\end{figure}

This use case is focused on providing end-users with a time-critical tool for requesting a UGV or UAV for specific missions, such as casualty extraction, medical assistance, or equipment delivery. With this feature, the first responder shares their location by using a smartphone when sending the request. The FCC then generates a viable path and transmits it to the requested robot. Finally, the robot autonomously navigates along the generated path to reach the end-user's location, as shown in Figure \ref{fig:requests-case}.

The participants in this scenario, in order of involvement, are the first responder, their smartphone, the FCC, and the requested robot. In this case, the robot assists with tasks such as transporting casualties or equipment, while the smartphone enables the first responder to issue requests through the communication system. The purpose of the FCC is to receive the end-user request, generate an effective route plan, and send the path to the requested robot. This sequence of actions illustrates the importance of the communication system, which facilitates coordination among the first responders, the FCC, and the requested robot. Efficient communication is crucial for the success of critical-time missions, guaranteeing that assistance is delivered promptly, effectively, and on-site when needed.

This use case was conceived as a cooperative training exercise with a combat medical unit of the Spanish Army (Tercio "Alejandro Farnesio" $4^o$ of the Spanish Legion). The casualty extraction and evacuation scenario has been performed in different versions across individual editions of JEMERG. In these realistic exercises, run on a one-shot basis, we have used the Rover J8 UGV with the capacity to transport two victims in two commercial stretchers in each run (see Figure \ref{fig:requests-case}-a). One of the features we have added to this cooperative mission is the integration of the UMA-ROS-Android application for the first responder requests and a path planner in the SAR-IoCA architecture \cite{Toscano-Moreno2022}. Although in the initial experiment, the UGV was teleoperated through the extraction route due to inadequate autonomous navigation speed, last year it successfully performed autonomous navigation at high speeds safely.

This experiment was also conducted in a different scenario, utilizing a remote control center (BCC) to autonomously deploy a UGV in an urban SAR scenario at a distance of 440 km \cite{Sanchez-Montero2023}. This experiment highlights the limitations of commercial 5G network availability in SAR operations and underscores the critical need for \emph{ad-hoc} networks to address communication challenges in disaster situations.

\subsection{SLAM through Multi-Access Edge Computing}

In conflict zones, communication infrastructures are frequently targeted and disabled deliberately, aiming to isolate affected areas. Similarly, natural disasters such as earthquakes, hurricanes, or floods often damage critical infrastructure, while extended power outages further accelerate the breakdown of communication systems. These failures significantly compromise emergency response efforts, especially in SAR missions, where reliable connectivity is vital.

To address these challenges, 5G network slicing offers a compelling approach by allowing the creation of virtual network segments configured to meet the specific requirements of emergency services in terms of latency, bandwidth, and reliability. This ensures that, even in the face of widespread infrastructure disruption or heavy network traffic, essential SAR operations can continue to function with reliable communication links.

\begin{figure}[t]
     \centering     \includegraphics[width=\columnwidth]{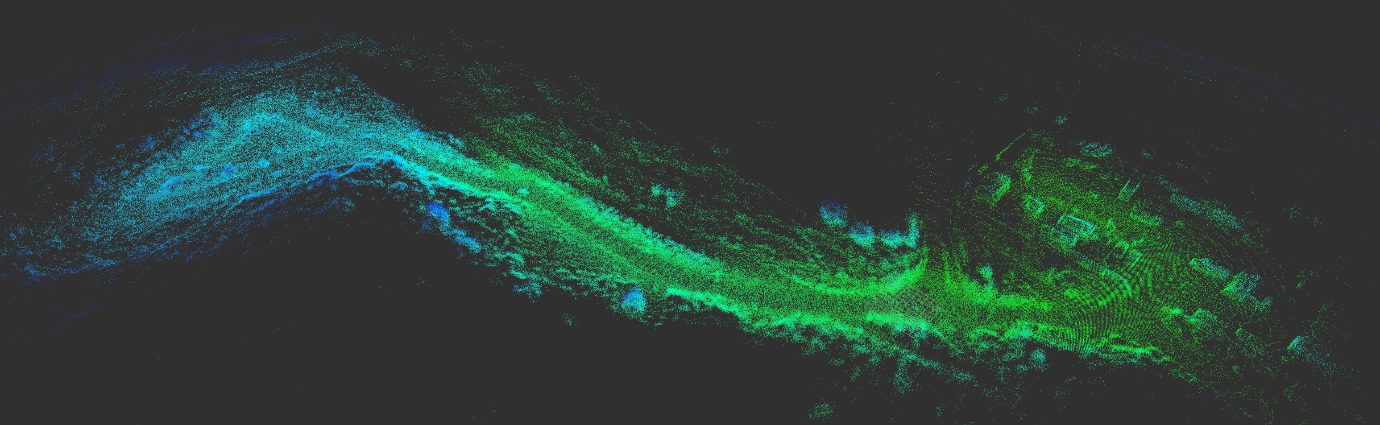}
     \caption{SLAM generated in a MEC center}
     \label{fig:slam-case}
\end{figure}

The implemented algorithm is the Hierarchical Dynamic Layers (HDL) Graph SLAM\footnote{\scriptsize https://github.com/koide3/hdl\_graph\_slam}, which supports 6 DOF real-time mapping. This algorithm, chosen due to its computational demands, benefits from Cloud computing when running on mid- or low-end hardware. Figure~\ref{fig:slam-case} shows how GNSS and LiDAR data from the Rover J8 are fused to generate a consistent map of the environment, clearly identifying key areas such as the CCCA.

The robot publishes map data at a rate of 6.58~MB/s ($\approx$53~Mbps), a throughput manageable by a 5G router connected to a non-commercial cell. Such data rates and concurrent sharing of network resources with other Edge devices would not be feasible without 5G and the use of network slicing, where a dedicated ROS 1 subnetwork is assigned to each network slice. As the map grows, message size increases, making 5G's low latency critical (below 19 ms in SA mode in these experiments) —something unachievable with LTE or 4G. 

Network slicing also ensures fair resource allocation within each slice, allowing Rover J8’s router SIM to handle additional tasks, such as high-frequency GNSS streaming or on-board camera video upload.

\subsection{Multi-robot cooperative exploration and casualty extraction in communication-denied areas}

This use case aimed to validate an implementation of the IoCA architecture in a SAR mission in order to test the whole multi-robot system using 5G communications and support real-time requirements, high bandwidth, and low latencies. These agents can include human responders such as firefighters, military units, and police, as well as robotic units like UGVs and UAVs, or canine rescue teams \cite{Fernandez-Lozano2018}.

The experimental validation of the SAR-IoCA architecture was performed during a realistic disaster response exercise, part of the full-scale exercise in the annual JEMERG event \cite{Bravo-Arrabal2021}. The simulated mission was conducted under realistic conditions following strict timing constraints and safety protocols coordinated by the exercise director and the organization staff.


    \begin{figure}[t]
    \centering
    \includegraphics[width=0.95\columnwidth]{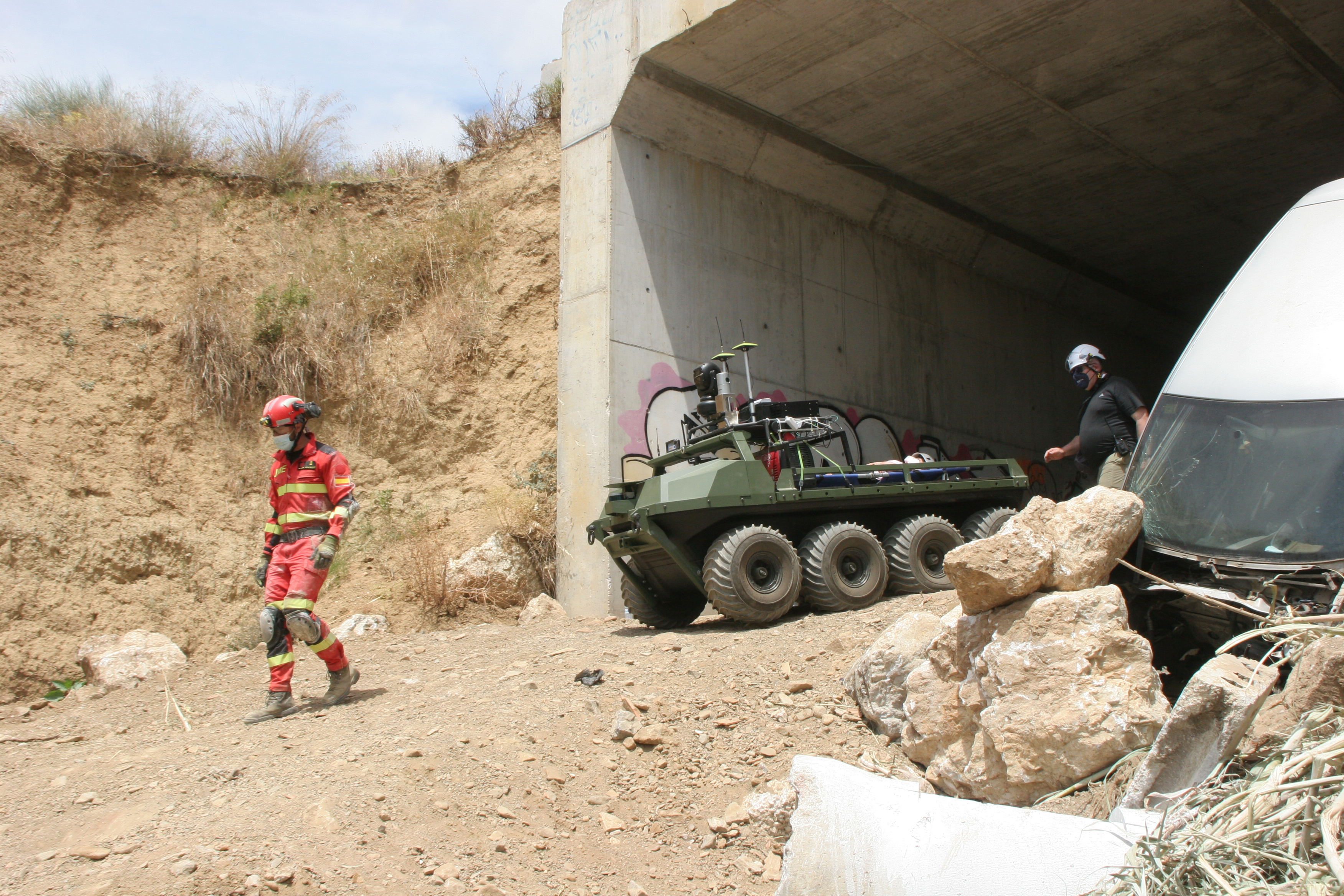}
    \caption{Rover J8 coming out of a tunnel in \textit{follow me} mode.}
    \label{fig:comm-denied-case}
\end{figure}


The mission was an emergency response to a disaster caused by an earthquake, which created fires and left victims trapped inside crushed vehicles in a storm-water drainage tunnel with challenging access. Other participants included the Provincial Fire Department of Málaga and the Spanish Military Emergency Unit (UME). The mission protocol was established in collaboration with the participant organizations to facilitate cooperation among heterogeneous robotic teams.

The use of the SAR-IoCA architecture facilitated the coordination of three UGVs and one UAV with the human rescue teams. The UAV was used to map the environment, creating a Digital Elevation Model (DEM) and an orthophoto of the operation area. These maps serve two purposes: to help the human rescue teams in assessing the disaster and to allow path planning for cooperative robotic agents. At the FCC, human coordinators are responsible for making decisions about the deployment of rescue and robotic teams based on the data received from the field through the different H-WSN and the streamed video via ROS. Both the UAV and UGVs used their onboard sensors to detect risks and potential victims. 

The Rover J8 UGV requires special attention due to its high bandwidth sensor data transmission needs. The vehicle streams 4K video from an onboard smartphone and LiDAR sensor point cloud data to the FCC. However, when the robot enters communication-denied areas—where even GNSS may fail—the SAR coordinator (located at the Forward Command Center, FCC) can no longer access the camera feeds, making it necessary for a UME responder (see Figure \ref{fig:comm-denied-case}) to press the \textit{follow me} button located on the front of the UGV. Additionally, the smartphone application can automatically reduce the resolution of the transmitted video or control when it is sent, acting as a redundant vision system when the robot's onboard systems fail. This highlights the usefulness of integrating a smartphone into ROS as a plug-and-play Edge device, independent from the robot's own systems.

A coordinated effort is then made to extract and evacuate casualties. This process of exploration and response to identified risks and victims continued as necessary until is required by the mission to ensure the safety of the emergency area. The mission was completed successfully, with victims found and evacuated in the designated area as a result of the joint operation of a fire brigade team, a military emergency team, and our systems (comprising the UGVs, UAV, FCC, and BCC). The experiments lasted one hour and fourteen minutes.

The specialist rescue teams involved in the realistic exercises were satisfied with the overall results. The MEC centers played a significant role by providing fluid audiovisual information, which was enhanced through aerial and ground exploration of the environment. This included the teleoperation of the UGVs and the effective identification of victims.

\section{Discussion}
\label{sec:discussion}

Emergency operations present unique challenges that increasingly benefit from robotic systems and advanced communications technology. The UMA's Robotics and Mechatronics Lab has pioneered an integrated approach that recognizes the fundamental importance of communications infrastructure in maximizing the effectiveness of robotic emergency response.
Emergency scenarios, by their nature, require coordinated efforts across multiple responders and systems. A cooperative approach to robotic systems, considering the use of heterogeneous specialized robots, promises a greater impact in the effectiveness of emergency missions.
This cooperative approach relies on the ability to share and exchange information among multiple robots, but also among the multiple agents (humans, crewed vehicles, command and control centers) needed to accomplish an actual mission. Communications are then a central element. Thus, an integrated approach must pay attention to communications just like to any other parts of the overall system, driving to a co-design process of robotics and communications.

The emergence of 5G has provided high bandwidth and low latency communications. One of the first benefits was an easier and improved teleoperation of ground and air robots, thanks to the availability of high-definition video for the operators. But more importantly, 5G has enabled new use cases within emergency operations. 5G networks facilitate the real-time transmission of large amounts of sensor data from robots (UGVs and UAVs). This includes data from onboard cameras for remote viewing in either forward or remote command and control centers, which effectively changes the situational awareness of mission coordinators. But the transmission of LiDAR point clouds is also possible, allowing that the processing can be moved out of the robot towards the Edge or even out of the field. Our experience has shown that running SLAM algorithms in a MEC center far away from the operation terrain can be similarly effective to local execution. Moving computation from the robot can benefit energy autonomy, but it allows also for the use of more complex algorithms, requiring more computation power. 
Another use case enabled by 5G is the remote request of robots by human first responders. Humans can be integrated into the systems through the use of 5G user equipment. Special-purpose apps have been designed to share terminal sensor data like video, audio, or positioning. Thus, human responders can request a robot for some specific mission, like casualty extraction. Their location is shared with the request, and a viable path is generated in the FCC and sent to the requested robot. Then, the robot follows the generated path autonomously to the extraction point. 

Multi-robot missions can be improved by the use of this approach, focusing on an active perception \cite{Queralta2020}. The FCC can receive information (video, audio, location, environmental data) from several robots and decide how to plan and execute the mission, including teleoperation and autonomous motion of the robots. 

The effective integration of all these heterogeneous systems, connected through a communication system, required specific efforts. An Internet of Cooperative Agents architecture (X-IoCA) \cite{Bravo-Arrabal2021} has been proposed as a framework to include heterogeneous sensor networks and robots, MEC, and advanced communications (including 5G) in field applications requiring the cooperation between human and robot teams. 

Being a central element in this cooperative approach, communications need to be reliable. The use of commercial communication systems has always been seen as a drawback by first responders. However, 5G has been received differently, because of two features: network slicing and crowdcells. Network slicing allows for private networks, isolating the traffic from certain users while granting performance. Crowdcells provide 5G communications at least at a local scale even when the commercial infrastructure has been disrupted.

\subsection{Lessons learned}

Through the comprehensive work of the Robotics and Mechatronics lab at the UMA, several valuable lessons have emerged regarding the effective deployment of robotic systems in emergency scenarios. 
First, applications have proven most effective when focusing on specific emergency scenarios rather than attempting to create universal solutions. This specialized approach has yielded particularly promising results in medical evacuation and reconnaissance operations, where robots can directly address well-defined operational challenges.
Emergency scenarios often require diverse robotic capabilities. The integration of both ground and aerial vehicles, along with specialized systems like robotic stretchers, provides emergency responders with a comprehensive toolkit adaptable to varying disaster environments, and to different needs.
Environmental information is key to situational awareness. WSNs can be useful to track information like temperature or gas levels, but also have shown promise for specific applications, like victim detection and localization in complex scenarios. But effectiveness is linked to a coordinated use. 
The human element remains central to effective emergency response. Robotic systems must be designed with careful consideration of operational procedures, responder workload, and human factors. The active involvement of fire departments, police, and other emergency organizations in the development process significantly improves adoption and operational effectiveness.
The X-IoCA architecture has demonstrated the critical importance of integrated communications frameworks that facilitate seamless data sharing between heterogeneous robots, sensor networks, and computing resources. This architectural approach enables coordinated action across multiple platforms in ways that isolated systems cannot achieve.
The integration of 5G technology offers transformative potential for emergency robotics, particularly through low latency communication and cloud robotics capabilities. However, realistic testing has revealed that communication challenges in emergency environments, such as signal propagation issues in confined spaces, require continuous refinement of these technologies for reliable field performance.
The co-design principle—simultaneously considering both robotic capabilities and communication network performance—has proven essential for developing effective systems. Traditional approaches that treat these as separate concerns inevitably lead to compromised performance in real-world deployments.
Software frameworks like ROS provide valuable integration capabilities, while custom applications facilitate critical human-robot interaction. Tools that enable responders to request robotic assistance from specific locations significantly enhance operational utility.

Compared to ROS 2, ROS 1 poses challenges for distributed SAR missions due to its centralized architecture and reliance on a single ROS master, limiting scalability, fault tolerance, and adaptability in dynamic, multi-robot scenarios. In contrast, ROS 2 adopts a decentralized communication model based on Data Distribution Service (DDS), which enables peer-to-peer discovery, simplifies configuration, and enhances resilience across heterogeneous networks. Its support for Quality of Service (QoS) ensures reliable communication over unstable links. In our experience, ROS 2 has also shown more stable behavior over Virtual Private Network (VPN) connections, further reinforcing its suitability for coordinating autonomous agents in disaster response.

The evolution toward systems of systems approaches reflects the recognition that emergency response requires coordinated action across multiple platforms rather than isolated robotic capabilities. This interconnected perspective enables more comprehensive and adaptable response strategies.
Finally, realistic field testing with emergency services has proven indispensable for identifying practical limitations and opportunities for improvement. 

\subsection{Challenges for the evolution towards 6G}

The feedback obtained through these exercises guides development far more effectively than laboratory testing alone, ensuring that technological innovations remain firmly grounded in operational realities.
These lessons learn to disclose some relevant challenges:
\begin{itemize}
    \item Reliability is critical, not only in robots but also in communications. Emergency operations often are safety-critical.   \item Sensing is still a limitation. A more effective integration of robots and humans in cooperative missions relies on having better information from the surroundings of the robot. The ability to perceive humans is also key in missions like victim detection. 
    \item Better human-robot interfacing. What human responders require from robots is frequently complex. Immersive interfaces, or natural language interfaces can be a solution, although they pose heavy requirements on communications and computing power. 
    \item Natural integration of robotics and communications. Many applications can benefit from connected robots, especially if both robots and communication systems are conceived for this integration. 
\end{itemize}

6G may help with these challenges. To improve reliability, 6G provides ultra-reliable low-latency communication (URLLC), ensuring stable connections for safe teleoperation and coordination of robots in hazardous environments \cite{one6g2025}. This helps prevent communication failures that could compromise mission success. 
For sensing, 6G deeply integrates wireless sensing with communication, expanding a robot’s awareness beyond its onboard sensors. By combining RF sensing, thermal imaging, and deep learning algorithms, robots can detect disaster-related elements \cite{Gonzalez-Navarro2023} and even identify human presence beneath debris by analyzing subtle movements or vital signs. High-bandwidth data transmission enables the creation of detailed, real-time environmental maps, improving navigation and situational awareness. Additionally, computational offloading allows robots to transfer data-heavy perception tasks to network-Edge computing, reducing their onboard processing load and enabling faster, more sophisticated analysis.
In terms of human-robot interaction (HRI), 6G’s high bandwidth and low latency improve telepresence, allowing operators to control robots more effectively through high-definition video and audio. The advanced Edge computing capabilities of 6G networks also support natural language processing, making robot interactions more intuitive and accessible for human responders.
Beyond these individual improvements, 6G is built with robotics in mind, facilitating seamless integration between robotic systems and communication networks. It enables predictive connectivity management, allowing robots to adapt their operations based on expected network conditions. Dynamic resource allocation ensures robots receive the appropriate bandwidth and reliability for their tasks. Additionally, Integrated Sensing and Communication (ISAC) turns the communication network itself into a sensing platform, supplementing robots with environmental data and reducing their dependence on onboard sensors. Lastly, 6G enhances interoperability, enabling efficient data sharing and coordination among diverse robotic teams and human responders, ensuring smoother collaboration in emergency scenarios.

\section{Conclusion}
\label{sec:concl}

This work outlines the work and lessons learned by the Robotics and Mechatronics Lab at the UMA in the field of emergency robotics, with a significant focus on the use of communications together with robotic systems. The successful deployment of advanced robotic systems, particularly as interconnected systems of systems, heavily relies on robust and reliable communication infrastructure. Our experience highlights the relevance of co-design of robotic and communication systems, and how effectiveness can be fostered by this approach. New use cases have been unlocked by the addition of 5G communications, but challenges still exist. The key takeaway is the potential of a holistic approach to robotics and communications in saving lives and mitigating the impact of emergencies.

\section*{Acknowledgements}
\noindent The authors thank the Chair for Safety, Emergencies and Disasters of the University of Malaga, led by Prof. Jesús Miranda-Páez for organizing the exercises. Finally, we wish to acknowledge the support from our colleagues of the UMA Robotics and Mechatronics Group. We would also like to thank all the organization staff and participants involved in the exercises. This work has been performed in the frame of the project ``SAR 4.0:  Leapfrogging to a New Paradigm in Cooperative Human-Robot Cyber-physical Systems for Search and Rescue'', funded by the Spanish Government (PID2021-122944OB-I00), and the “Piloto 5G Andalucía” initiative, project C007/18-SP, promoted by the Ministerio de Asuntos Económicos y Transformación Digital, through Red.es, and co-financed by the European Regional Development Fund, being developed by Vodafone and Huawei.

\bibliographystyle{ieeetr}
\bibliography{bibliography}

\end{document}